\documentclass[letterpaper]{article} % DO NOT CHANGE THIS
\usepackage[submission]{aaai2026}  % DO NOT CHANGE THIS
\usepackage{times}  % DO NOT CHANGE THIS
\usepackage{helvet}  % DO NOT CHANGE THIS
\usepackage{courier}  % DO NOT CHANGE THIS
\usepackage[hyphens]{url}  % DO NOT CHANGE THIS
\usepackage{graphicx} % DO NOT CHANGE THIS
\usepackage{natbib}  % DO NOT CHANGE THIS AND DO NOT ADD ANY OPTIONS TO IT
\usepackage{caption} % DO NOT CHANGE THIS AND DO NOT ADD ANY OPTIONS TO IT
\usepackage{algorithm}
\usepackage{algorithmic}

\usepackage{newfloat}
\usepackage{listings}
\DeclareCaptionStyle{ruled}{labelfont=normalfont,labelsep=colon,strut=off} % DO NOT CHANGE THIS
\floatstyle{ruled}
\newfloat{listing}{tb}{lst}{}
\floatname{listing}{Listing}
\usepackage{amsmath}
\usepackage{amsfonts}
\usepackage{graphicx}
\usepackage{tikz}
\usetikzlibrary{positioning, shapes, arrows.meta}
\definecolor{myblue}{RGB}{0,102,204}
\usepackage{subcaption} 
\usepackage{booktabs}

\makeatletter
\def\@copyrighttext{}
\def\@numberofauthors{0}
\def\@version{}
\makeatother

\makeatletter
\renewcommand{\@maketitle}{}
\renewcommand{\@author}{}
\renewcommand{\@title}{}
\makeatother

\begin{document}

\twocolumn[{  % <-- Open single-column block

\begin{center}
    {\Large \textbf{BinConv: A Neural Architecture for Ordinal Encoding in Time-Series Forecasting}}\\[1.5ex]
    \textbf{Andrei Chernov\textsuperscript{1}, Vitaliy Pozdnyakov\textsuperscript{2}, Ilya Makarov\textsuperscript{3}}\\
    \textsuperscript{1}Independent Researcher; Germany\\
    \textsuperscript{2}AIRI, HSE University; Russia\\
    \textsuperscript{3}AIRI, ISP RAS, ITMO University; Russia\\
    \texttt{chernov.andrey.998@gmail.com,  vvpozdnyakov@hse.ru, makarov@airi.net}\\[2ex]
\end{center}

}]  % <-- Close single-column block and resume 2-column layout

% \author{%
%   Andrei Chernov \\
%   % \thanks{Use footnote for providing further information
%   %   about author (webpage, alternative address)---\emph{not} for acknowledging
%   %   funding agencies.} \\
%   Independent Researcher \\
%   Munich, Germany \\
%   % Cranberry-Lemon University\\
%   % Pittsburgh, PA 15213 \\
%   \texttt{chernov.andrey.998@gmail.com} \\
%   % examples of more authors
%   \And
%   Vitaliy Pozdnyakov \\
%   AIRI \\
%   Moscow, Russia \\
%   \texttt{pozdnyakov@airi.net} \\
%   \AND
%   Ilya Makarov \\
%   AIRI, ISP RAS, ITMO University \\
%   Moscow, Russia \\
%   \texttt{makarov@airi.net} \\
%   % \And
%   % Coauthor \\
%   % Affiliation \\
%   % Address \\
%   % \texttt{email} \\
%   % \And
%   % Coauthor \\
%   % Affiliation \\
%   % Address \\
%   % \texttt{email} \\
% }

% REMOVE THIS: bibentry
% This is only needed to show inline citations in the guidelines document. You should not need it and can safely delete it.
% \usepackage{bibentry}
% END REMOVE bibentry

% \begin{document}

% \maketitle

\begin{abstract}
Recent work in time series forecasting has explored reformulating regression as a classification task. By discretizing the continuous target space into bins and predicting over a fixed set of classes, these approaches benefit from more stable training, improved uncertainty modeling, and compatibility with modern deep learning architectures. However, most existing methods rely on one-hot encoding, which ignores the inherent ordinal structure of the target values. As a result, they fail to convey information about the relative distance between predicted and true values during training. In this paper, we address this limitation by applying \textbf{Cumulative Binary Encoding} (CBE), a monotonic binary representation that transforms both model inputs and outputs. CBE implicitly preserves ordinal and magnitude information, allowing models to learn distance aware representations while operating within a classification framework. To leverage CBE effectively, we propose \textbf{BinConv}, a fully convolutional neural network architecture designed for probabilistic forecasting. We demonstrate that standard fully connected layers are not only less computationally efficient than convolutional layers when used with CBE, but also degrade forecasting performance. Our experiments on standard benchmark datasets show that BinConv achieves superior performance compared to widely used baselines in both point and probabilistic forecasting, while requiring fewer parameters and enabling faster training.

\end{abstract}


\section{Introduction}
\label{sec:introduction}
Time series forecasting plays a key role in decision-making across domains such as finance \cite{akita2016deep}, healthcare \cite{song2020distributed}, retail \cite{bandara2019sales}, and climate science \cite{zaini2022systematic}. While classical methods like ARIMA and Exponential Smoothing remain widely used in practice \cite{hyndman2018forecasting}, recent advances increasingly favor deep learning approaches \cite{lim2021time,li2024deep}, particularly in scenarios requiring high-precision forecasting \cite{zhang2024probts}.

A critical component of deep learning-based forecasting is data preprocessing. This typically includes normalization, min-max or mean scaling \cite{salinas2020deepar}, Box-Cox transformation \cite{hyndman2018forecasting}, and short-time Fourier transforms \cite{furman2023short}. Recent studies \cite{ansari2024chronos,liu2024timer} have explored discretizing the continuous target space into fixed-size bins. Values are then represented as one-hot vectors, enabling a tokenized view of time series aligned with Transformer-style architectures. Empirical results suggest that such representations enhance performance, especially within Time Series Foundation Models \cite{zhang2024probts}.

However, one-hot encoding lacks ordinal awareness, treating bins as unrelated categories and ignoring relative distances, which are important for modeling temporal dynamics. To address this, we propose Cumulative Binary Encoding (CBE): a representation where all bins below (or equal to) the current value are activated. This monotonic encoding, previously applied to tabular data \cite{gorishniy2022embeddings}, preserves global ordinal structure and yields richer inputs for neural models. % Figure~\ref{fig:CBE_example} shows an example of CBE applied to a time series.

%\begin{figure*}[ht]
%    \centering
%    \begin{subfigure}{0.3\textwidth}
%        \includegraphics[width=\linewidth]{images/time_series.pdf}
%    \end{subfigure}
%    \begin{subfigure}{0.3\textwidth}
%        \includegraphics[width=\linewidth]{images/one_hot.pdf}
%    \end{subfigure}
%    \begin{subfigure}{0.3\textwidth}
%        \includegraphics[width=\linewidth]{images/CBE.pdf}
%    \end{subfigure}
%
%\caption{Visualization of different representations of time series data. \textbf{(left)} Original time series showing monthly airline passenger counts. \textbf{(center)} One-hot encoding of discretized values. \textbf{(right)} Cumulative Binary Encoding (CBE), where all bins less than or equal to the active bin are highlighted, preserving ordinal structure.}
%\label{fig:CBE_example}
%\end{figure*}

Choosing an appropriate neural architecture is another central design decision. RNNs, CNNs, and Transformers are commonly used, with Transformer based models \cite{zhang2024probts} excelling due to their capacity for long range temporal modeling \cite{vaswani2017attention}. Yet, their nearly quadratic complexity with sequence length \cite{tay2022efficient} poses practical challenges. 
%Most importantly, in Section~\ref{sec:ablation_study}, we show that using fully connected layers with CBE leads to performance degradation and provide an explanation for this behavior. Since fully connected layers are the core components of Transformer and RNN architectures, these models are not well suited for time series forecasting with CBE, leaving us with convolutional architectures as the most viable option.

As a result, in this paper we introduce \textbf{BinConv}, a CNN-based model that combines two dimensional and one dimensional convolutions to process CBE representations efficiently. Moreover, BinConv with CBE naturally supports probabilistic output by modeling distributions over bins, a capability often missing in standard CNN forecasting models.  Remarkably, CBE and BinConv are effective only when used together. Our ablation study in Section~\ref{sec:ablation_study} shows that neither CBE combined with a Transformer model nor one-hot encoding used with BinConv achieves satisfactory performance.

% The main current limitation of our architecture is that it does not naturally support multivariate time series forecasting. Therefore, to forecast multivariate time series, one must forecast each series independently. Nevertheless, to enable a fair comparison, we evaluate our method on three univariate time series datasets and five multivariate time series datasets from the ProbTS benchmark \cite{zhang2024probts}, which includes diverse datasets from multiple domains. Our model consistently outperforms strong baselines in both point and probabilistic forecasting tasks for the univariate benchmarks and, surprisingly, performs no worse on the multivariate datasets, despite the current lack of architectural support for modeling multivariate structure.

% Finally, we conduct an ablation study to demonstrate that CBE and BinConv are effective only when used together. Specifically, we show that neither CBE with transformer architectures nor one-hot encoding with BinConv yields good performance. 
% The main contributions of this work are:
% \begin{itemize}
%   \item A novel discrete representation, \textbf{Cumulative Binary Encoding (CBE)}, which captures ordinal and metric structure in time series.
%   \item A lightweight convolutional model, \textbf{BinConv}, combining 2D and 1D convolutions to process CBE sequences efficiently.
%   \item Extensive evaluation on the ProbTS benchmark, showing that our method achieves state-of-the-art results across multiple forecasting datasets.
% \end{itemize}

\section{Related Work}
\label{sec:rw}

%\textbf{Classical and Deep Learning Methods for Time Series Forecasting.}
%Classical models such as ARIMA, exponential smoothing, and vector autoregression are widely used for their simplicity and efficiency \cite{hyndman2018forecasting}. However, they often underperform on large-scale datasets exhibiting complex and nonlinear patterns, as shown in M4, M5, M6 forecasting competitions \cite{makridakis2018m4,makridakis2022m5,makridakis2024m6}. Deep learning approaches improve upon these by framing forecasting as a sequence-to-sequence problem using fixed context windows \cite{lim2021time}. Recent developments like Time Series Foundation Models \cite{liang2024foundation,das2024decoder,darlow2024dam} leverage large-scale pretraining to generalize across domains, achieving strong results in both point and probabilistic forecasting \cite{zhang2024probts}.

\textbf{Neural Network Architectures for Time Series Forecasting.}
Neural models vary in their trade-offs between accuracy, scalability, and speed. Transformer-based models such as PatchTST \cite{nie2023a}, TFT \cite{lim2021temporal}, and Informer \cite{zhou2021informer} capture long-range dependencies well, but incur high computational costs due to their quadratic scaling with sequence length. Earlier RNN-based models like LSTM \cite{hochreiter1997long}, GRU \cite{che2018recurrent}, and DeepAR \cite{salinas2020deepar} are more efficient at inference, though limited by their sequential nature, which hinders training parallelism. CNN-based architectures such as TCN \cite{bai2018empirical}, DeepTCN \cite{chen2020probabilistic}, and ModernTCN \cite{luo2024moderntcn} offer fast, parallel training and strong forecasting performance. Following this line, our CNN-based BinConv model balances accuracy and efficiency when used with CBE inputs.

\textbf{Token-Based Representations of Time Series.}
Inspired by NLP, recent work reformulates time series forecasting as next-token prediction using discretization. Models like LLMTime \cite{gruver2023large}, TiMER \cite{liu2024timer}, and Chronos \cite{ansari2024chronos} encode values as digits, patches, or bins. Gorishniy et al.~\cite{gorishniy2022embeddings} proposed piecewise linear encoding (PLE) for tabular data, preserving ordinal structure. Most models rely on one-hot encoding, which discards distance and order information between bins. This limits the ability to learn value relationships. In contrast, our CBE captures both order and magnitude, improving learning while remaining compatible with token-based models.

%\textbf{Probabilistic Forecasting with Deep Learning Methods.}
%Probabilistic forecasting offers both predictions and uncertainty. Deep learning models approach this via quantile regression (e.g., TFT, DeepAR), parametric output distributions (e.g., DeepTCN), or generative models like TimeGrad \cite{rasul2021autoregressive} and GRU-NVP \cite{rasul2021multivariate}. While effective, generative approaches can be unstable and costly to train. Token-based models such as Chronos and TiMER enable probabilistic inference via sampling in discrete space. Our model adopts this approach using CBE, supporting both point and probabilistic forecasts through structured sampling.

\section{Cumulative Binary Encoding}
\label{sec:CBE}
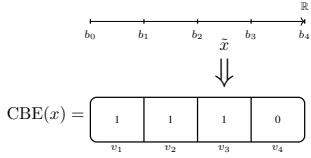
\begin{figure}[ht]
\centering
\resizebox{0.5\linewidth}{!}{%
\begin{tikzpicture}[baseline, every node/.style={font=\scriptsize}]
% Parameters
\def\boxwidth{1.2}
\def\nbins{4} % Number of embedding values

% Axis line (ends at b4)
\draw[->] (0,0) -- ({\boxwidth*\nbins},0);
\foreach \i in {0,...,4} {
    \pgfmathsetmacro{\x}{\boxwidth*\i}
    \draw[thick] (\x,0.05) -- (\x,-0.05);
    \node at (\x, -0.3) {$b_{\i}$};
}

% Real line label
\node at ({\boxwidth*\nbins}, 0.35) {$\mathbb{R}$};

% x and arrow (centered between b2 and b3)
\pgfmathsetmacro{\xmid}{\boxwidth*2.5}
\node at (\xmid, -0.5) {\large$\Tilde{x}$};
% \draw[very thick, ->] (\xmid, -1) -- (\xmid, -1.4);
\node at (\xmid, -1.15) {\Huge$\Downarrow$};

% CBE label
\node at ({-1.0}, -2.1) {\large$\mathrm{CBE}(x) =$};

% Embedding box (ends at b4)
\draw[thick, rounded corners=4pt] (0, -1.7) rectangle ({\boxwidth*\nbins}, -2.7);
\foreach \i/\val in {1/1, 2/1, 3/1} {
    \pgfmathsetmacro{\x}{\boxwidth*\i}
    \draw[thick] (\x, -1.7) -- (\x, -2.7);
}
% Fill values
\foreach \i/\val in {0/1, 1/1, 2/1, 3/0} {
    \pgfmathsetmacro{\x}{\boxwidth*\i}
    \node at (\x + \boxwidth/2, -2.2) {\val};
    \node at (\x + \boxwidth/2, -2.85) {$v_{\the\numexpr\i+1}$};
}
\end{tikzpicture}
}
\vspace{-1ex}
\caption{Example of CBE encoding for 4 bins.}
\label{fig:CBE}
\end{figure}

The scale of time series can vary significantly, even within a single dataset. Therefore, before converting time series values into an encoding space, we apply normalization to map values into a quantized range. Similar to the Chronos model \cite{ansari2024chronos}, we use the mean scaling technique \cite{salinas2020deepar}, where each value is divided by the mean over a given context window. Specifically, we compute the scaled value as $\Tilde{x} = x / s$, where $s = \frac{1}{C} \sum_{i=1}^{C} \left| x_i \right|$.

Next, we quantize the scaled time series as follows: 
\begin{align}
\mathrm{CBE}(x) &= [v_1, \dots, v_{D}] \in \mathbb{R}^D \notag \\
v_d &= 
\begin{cases}
0, & \Tilde{x} < b_d \\
1, & \Tilde{x} \geq b_d
\end{cases},
\label{eq:CBE}
\end{align}
where $b_d$ indicates a $d$-th bin in the original space. An example of construction of CBE is shown in Figure~\ref{fig:CBE}.

The interval from $b_0$ to $b_D$ is uniformly split into $D$ bins in a data-agnostic manner, where $D$, $b_0$, and $b_D$ are hyperparameters. In our experiments, we do not fine-tune these values per dataset and use fixed settings of $D = 1000$, $b_0 = -5$, and $b_D = 5$, unless stated otherwise. This quantization results in Cumulative Binary Encoding of the form $111\ldots100\ldots0$.

This type of encoding is not new in the context of tabular deep learning for numerical features and can be viewed as a simplified version of the piecewise linear encoding proposed in \cite{gorishniy2022embeddings}. However, unlike previous work, we not only transform the inputs using this scheme, but also forecast targets directly in the encoding space, as detailed in Section~\ref{sec:forecasting}. After generating a forecast in the encoding space, we apply an inverse transformation by assigning the average value of the bin encoding, i.e., $\Tilde{y} = (b_i + b_{i+1}) / 2$, where $i$ is the index of the last $1$. Finally, to obtain the forecast in the original space, we reverse the mean scaling: $y = s \cdot \Tilde{y}$. To the best of our knowledge, this approach is novel in the context of time series forecasting.

\section{BinConv}

\subsection{Architecture}
\label{sec:BinConv_architecture}

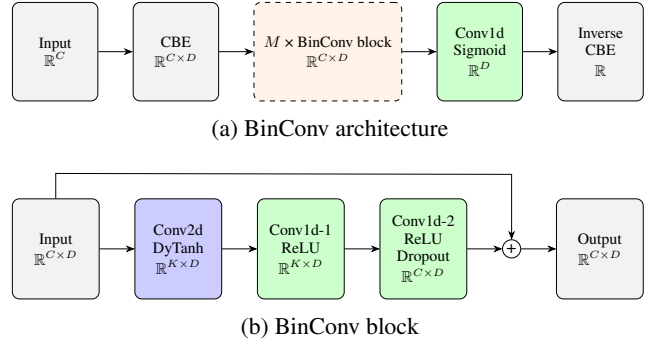
\begin{figure}[t]
\centering

\begin{subfigure}{\columnwidth}
\centering
\resizebox{\columnwidth}{!}{%
\begin{tikzpicture}[
    node distance=1.5cm and 0.7cm,
    line width=0.5pt,
    >=Stealth,
    every path/.style={->},
    block/.style={
      rectangle, draw, rounded corners,
      text width=1.5cm, align=center, minimum height=2cm,
      font=\footnotesize, fill=gray!10
    },
     greenblock/.style={
      rectangle, draw, rounded corners,
      text width=1.5cm, align=center, minimum height=2cm,
      font=\footnotesize, fill=green!20
    },
    repeatblock/.style={
      rectangle, draw, dashed, rounded corners,
      align=center, minimum height=2cm,
      minimum width=3.0cm,
      font=\footnotesize, fill=orange!10
    }
  ]
  \node[block] (input) {Input \\ $\mathbb{R}^{C}$};
  \node[block, right=of input] (CBE) {CBE \\[2pt] $\mathbb{R}^{C \times D}$};
  \node[repeatblock, right=of CBE] (binconv) {$M$ × BinConv block \\[2pt] $\mathbb{R}^{C \times D}$};
  \node[greenblock, right=of binconv] (finalconv) {Conv1d \\ Sigmoid \\[2pt] $\mathbb{R}^{D}$};
  \node[block, right=of finalconv] (output) {Inverse CBE \\[2pt] $\mathbb{R}$};

  \draw[->] (input) -- (CBE);
  \draw[->] (CBE) -- (binconv);
  \draw[->] (binconv) -- (finalconv);
  \draw[->] (finalconv) -- (output);
\end{tikzpicture}
}
\subcaption{BinConv architecture}
\label{fig:arch}
\end{subfigure}

\vspace{1em}

% --- Subfigure (b) ---
\begin{subfigure}{\columnwidth}
\centering
\resizebox{\columnwidth}{!}{%
\begin{tikzpicture}[
    node distance=1.5cm and 0.7cm,
    >=Stealth,
    block/.style={
      rectangle, draw, rounded corners,
      text width=1.5cm, align=center, minimum height=2cm,
      font=\footnotesize, fill=gray!10
    },
    blueblock/.style={block, fill=blue!20},
    greenblock/.style={block, fill=green!20},
    annot/.style={font=\footnotesize, align=center}
  ]
  \node (in)  [block]    {Input \\[2pt] $\mathbb{R}^{C\times D}$};
  \node (dw)  [blueblock,right=of in]  {Conv2d \\ DyTanh \\[2pt] $\mathbb{R}^{K\times D}$};
  \node (ff1) [greenblock,right=of dw] {Conv1d-1  \\ ReLU \\[2pt] $\mathbb{R}^{K\times D}$};
  \node (ff2) [greenblock,right=of ff1]{Conv1d-2 \\ ReLU \\ Dropout \\[2pt] $\mathbb{R}^{C\times D}$};
  \node (sum) [circle, draw, right=of ff2, inner sep=1pt] {+};
  \node (out) [block, right=of sum]    {Output\\[2pt] $\mathbb{R}^{C\times D}$};

  \draw[->] (in)  -- (dw);
  \draw[->] (dw)  -- (ff1);
  \draw[->] (ff1) -- (ff2);
  \draw[->] (ff2) -- (sum);
  \draw[->] (sum) -- (out);

  \draw[->]
    (in.north) -- ++(0,5mm) -- 
    ([yshift=13mm]sum.north) -- 
    (sum.north);
\end{tikzpicture}
}
\subcaption{BinConv block}
\label{fig:binconv}
\end{subfigure}

\caption{BinConv architecture. (a) First, we preprocess the time series using CBE. Then, the BinConv model is trained in the CBE space, and during forecasting, the inverse CBE transformation is applied to obtain predictions in the original space. BinConv consists of $M$ stacked BinConv blocks followed by a final 1D convolution layer with a large kernel size. (b) Each BinConv block is the core component of the architecture, comprising three convolutional layers and a residual connection. The DyTanh activation function is used as a substitute for layer normalization. The output dimension of each BinConv block matches its input dimension.}

\label{fig:binconv_full}
\end{figure}

\begin{table}[ht]
\centering
\caption{Convolutional parameters used in BinConv.}
\label{tab:binconv-params}
\begin{tabular}{lccc}
\toprule
\textbf{Layer} & \textbf{Kernel Size} & \textbf{In/Out Channels} & \textbf{Groups} \\
\midrule
Conv2d     & $(C,\; s_1)$ & $1\ /\ K$  & $1$ \\
Conv1d-1   & $s_2$       & $K\ /\ K$  & $K$ \\
Conv1d-2   & $s_2$       & $K\ /\ C$  & $K$ \\
Conv1d     & $s_3$      & $C\ /\ 1$  & $1$ \\
\bottomrule
\end{tabular}
\end{table}

To model time series effectively in the discrete CBE space, the architecture must preserve ordinal relationships, support autoregressive forecasting, and maintain computational efficiency. To this end, we propose BinConv, a convolutional neural architecture that efficiently captures the local temporal and monotonic structure of CBE vectors by combining 2D and 1D convolutions.

The BinConv model consists of $M$ convolutional blocks followed by a final convolutional layer with a sigmoid activation function, producing an output vector in the CBE space (see Figure~\ref{fig:arch}). For each convolution, we add padding on both sides to preserve output dimensionality. To preserve the cumulative structure of the CBE representation, we use ones for left padding and zeros for right padding. % Notably, we do not use fully connected layers in the model. In fact, including fully connected layers degrades performance, as further discussed in Section~\ref{sec:fc_binconv}.

Each BinConv block comprises three convolutional layers (see Figure~\ref{fig:binconv}). The first is a 2D convolution with kernel size $(C, 3)$, where $C$ corresponds to the context length. This layer transforms the input matrix--obtained after applying CBE--into $K$ one-dimensional vector representations, where $K$ is the number of output channels and treated as a hyperparameter. Following the work of \cite{zhu2025transformers}, we use the DyTanh activation function, which replaces layer normalization in our architecture.

Next, we apply two 1D convolutional layers with ReLU activation, followed by dropout after the second layer. Inspired by \cite{liu2022convnet,luo2024moderntcn}, we employ depthwise convolution which is a special case
of grouped convolution where the number of groups equals the number of channels $K$. This significantly reduces computational cost without sacrificing performance (see Section~\ref{sec:dwconv}). Each BinConv block is designed such that the input and output dimensions are identical. This allows the use of a residual connection as the final operation in the block, enabling the stacking of multiple BinConv blocks while mitigating the vanishing gradient problem. The main parameters of the model are listed in Table \ref{tab:binconv-params}.

To train the model, we apply the binary cross-entropy loss to the model output. This approach is analogous to a multi-label classification setup, where each class is predicted independently using a "one-versus-all" scheme.

\subsection{Forecasting}
\label{sec:forecasting}
To forecast time series, we apply an autoregressive approach: forecasting at time step $t+1$, appending the result to the context, then forecasting at $t+2$, and so on. To do this, we need to transform the vector of sigmoid-activated outputs into a CBE. We interpret each component of the output as the success probability of an independent Bernoulli trial. Due to the design of the CBE, only sequences of the form $11\ldots10\ldots0$ are considered valid. Therefore, we normalize probability of each valid sequence  by the total probability of all valid sequences, denoted as $Z$. 

More formally, the probability of a valid encoding vector $e_m$, where the first $m$ components are 1 and the remaining $D - m$ components are 0, is given by:
\begin{equation}
\label{eq:prob}
    p(e_m) = \frac{\hat{p}(e_m)}{Z} = \frac{
        \left( \prod_{i=1}^{m} p_i \right) \left( \prod_{i=m+1}^{D} (1 - p_i) \right)
    }{
        \sum_{j=0}^{D} \left( \prod_{i=1}^{j} p_i \prod_{i=j+1}^{D} (1 - p_i) \right)
    },
\end{equation}
where $p_i$ is the sigmoid output at position $i$.

Given these probabilities, we can either sample $n$ trajectories autoregressively to obtain a probabilistic forecast, or take the argmax at each time step, which is faster but results in a point forecast. In Section~\ref{section:experiments}, we show that the point estimates of the point and probabilistic forecasts do not differ significantly, and both outperform the point estimates of other models.

\begin{figure}[ht]
\centering
\begin{tikzpicture}[
    font=\small,
    box/.style={draw, minimum size=0.8cm, align=center},
    one/.style={fill=gray!40},
    zero/.style={fill=white},
    arrow/.style={->, thick, >=Stealth},
    prob/.style={font=\scriptsize}
]

% Sigmoid outputs
\node at (0, 2.2) {Sigmoid outputs};

\node[box, fill=blue!10] at (0, 1.5) {$p_1 = 0.4$};
\node[box, fill=blue!10] at (0, 0.7) {$p_2 = 0.9$};
\node[box, fill=blue!10] at (0, -0.1) {$p_3 = 0.2$};

% Keep the original vertical arrow
\node at (1.2, 0.7) {\rotatebox[origin=c]{90}{{\Huge$\Downarrow$}}};

% Embedding 0: [0,0,0]
\node[box, zero] at (2.1, 2) {0};
\node[box, zero] at (2.8, 2) {0};
\node[box, zero] at (3.5, 2) {0};
\node[prob] at (5.4, 2) {\shortstack[l]{
$\hat{p}(e_0) = 0.048$ \\
$p(e_0) = \frac{0.048}{Z} \approx 0.109$
}};

% Embedding 1: [1,0,0]
\node[box, one] at (2.1, 1.2) {1};
\node[box, zero] at (2.8, 1.2) {0};
\node[box, zero] at (3.5, 1.2) {0};
\node[prob] at (5.4, 1.2) {\shortstack[l]{
$\hat{p}(e_1) = 0.032$ \\
$p(e_1) = \frac{0.032}{Z} \approx 0.073$
}};

% Embedding 2: [1,1,0]
\node[box, one, thick] at (2.1, 0.4) {1};
\node[box, one, thick] at (2.8, 0.4) {1};
\node[box, zero, thick] at (3.5, 0.4) {0};
\node[prob] at (5.4, 0.4) {\shortstack[l]{
$\hat{p}(e_2) \approx 0.288$ \\
$p(e_2) = \frac{0.288}{Z} \approx \mathbf{0.654}$
}};

% Embedding 3: [1,1,1]
\node[box, one] at (2.1, -0.4) {1};
\node[box, one] at (2.8, -0.4) {1};
\node[box, one] at (3.5, -0.4) {1};
\node[prob] at (5.4, -0.4) {\shortstack[l]{
$\hat{p}(e_3) = 0.072$ \\
$p(e_3) = \frac{0.072}{Z} \approx 0.164$
}};

% Legend box for Z
\node[draw=none, align=left] at (3.0, -1.3) {
$Z = \sum_m \hat{p}(e_m) = 0.048 + 0.032 + 0.288 + 0.072 = 0.44$
};

\end{tikzpicture}
\caption{Converting the neural network output $[0.4, 0.9, 0.2]$ into a CBE, the most probable sequence is $[1, 1, 0]$.}
\label{fig:CBE-left-to-right}
\end{figure}

\section{Experiments}
\label{section:experiments}

\subsection{Setup}
We evaluate performance on three univariate datasets: M4 Daily (daily), M4 Weekly (weekly), and Tourism Monthly (tourism) \cite{makridakis2020m4,athanasopoulos2011tourism}. The prediction horizon $H$ for each dataset is set to $14$, $13$, and $24$, respectively. The context length is set to $C = 3H$ for all datasets.

Additionally, we report long-term forecasting results on several multivariate datasets commonly used in the research community: ETTh1, ETTh2, ETTm1, ETTm2, and Illness \cite{zhang2024probts}. The context length $C$ and prediction horizon $H$ were set to $C = H = 36$ for the Illness dataset and $C = H = 96$ for the others.

For the baseline models, we used the optimal hyperparameters provided in~\cite{zhang2024probts}. In contrast, BinConv was evaluated using its default configuration without any fine-tuning. This configuration is consistent across univariate and multivariate datasets, with the exception of the number of bins (see Table~\ref{tab:binconv-hyperparams}). For multivariate datasets, we reduced the number of bins to accelerate experiment runtime.

We adopt the training and evaluation setup from \cite{zhang2024probts}. Specifically, we train each model for $50$ epochs, and report evaluation metrics on the test set. For all models, we used the Adam optimizer~\cite{kingma2015adam} for training. 

We evaluate point forecasts using the normalized mean absolute error (NMAE) and probabilistic forecasts using the continuous ranked probability score (CRPS), both widely adopted in time series forecasting research \cite{zhang2024probts, chen2020probabilistic, ansari2024chronos, rasul2021multivariate}. The NMAE is defined as
\[
\mathrm{NMAE} = \frac{\sum_{k=1}^K \sum_{t=1}^T \left| x_t^k - \hat{x}_t^k \right|}{\sum_{k=1}^K \sum_{t=1}^T \left| x_t^k \right|},
\]
where $K$ is the dimensionality of the time series ($K=1$ in the univariate case) and $T$ is the prediction horizon. The CRPS is given by
\[
\operatorname{CRPS}(F^{-1}, x) = \int_0^1 2 \Lambda_\alpha\left(F^{-1}(\alpha), x\right) \, d\alpha,
\]
with quantile loss $\Lambda_\alpha(q, z) = (\alpha - \mathbb{I}\{z < q\})(z - q)$ and indicator function $\mathbb{I}\{\cdot\}$. We compute CRPS using $19$ quantile levels and $100$ samples to assess the quality of probabilistic forecasts, and NMAE to evaluate point forecasting performance following to \cite{zhang2024probts}.

For BinConv, as discussed in Section~\ref{sec:CBE}, we applied mean scaling independently to each sample prior to quantization. This was necessary because certain datasets, such as the M4 benchmark, consist of time series with widely varying scales. %We did not fine-tune the BinConv model’s hyperparameters for each dataset, as it already demonstrated strong performance with the default configuration.

All computations were performed on NVIDIA Tesla A10 GPU with 22.5 GB of RAM using CentOS 7 operating system, and all models were implemented using PyTorch 2.7.

\subsection{Baseline Models}

%It is prohibitively time consuming to compare our approach with every existing model. Therefore, we select four representative baselines.

We select four representative baselines. First, we include PatchTST~\cite{nie2023a}, a transformer based model that achieves the best average performance across datasets in the ProbTS benchmark (Table 10 in~\cite{zhang2024probts}). Second, we evaluate DLinear~\cite{zeng2023transformers}, an efficient model in terms of both runtime and parameter count, while still achieving competitive performance.

Unlike BinConv, PatchTST and DLinear are non-autoregressive models: they predict the entire forecast horizon in a single inference pass. This design enables faster inference but lacks flexibility, as each change in forecast horizon requires retraining. In contrast, autoregressive models can adapt to arbitrary horizons at inference time. Additionally, PatchTST and DLinear are not probabilistic models and do not provide uncertainty estimates.

To address this gap in baselines, we include two autoregressive probabilistic models: GRU-NVP~\cite{rasul2021multivariate}, which combines a GRU encoder with normalizing flows; and TimeGrad~\cite{rasul2021autoregressive}, which applies score based diffusion modeling to generate future trajectories by denoising noise conditioned on past observations.

For all four baselines, we adapt the hyperparameter configurations provided by the ProbTS benchmark. 
%We do not tune the hyperparameters for BinConv, as it performs well with the default values, which are summarized for both univariate and multivariate datasets in Table~\ref{tab:binconv-hyperparams}. 

We exclude foundation time series models from our baselines, as their zero shot performance remains inferior to that of state-of-the-art local models~\cite{zhang2024probts, toner2025performance}.

%Overall, we argue that these four baselines are sufficient to evaluate the performance of BinConv. Furthermore, since our experimental setup exactly replicates that of the ProbTS benchmark, the results of BinConv can be directly compared to the full set of models reported therein.

\subsection{Results}
\subsubsection{Univariate Datasets}

\begin{table}[ht]
\centering
\caption{
Default hyperparameters used for BinConv. $C$ denotes the context length. The only difference between univariate and multivariate setups is the number of bins.
}
\label{tab:binconv-hyperparams}
\begin{tabular}{lcc}
\toprule
\textbf{Hyperparameter} & \textbf{Univariate} & \textbf{Multivariate} \\
\midrule
Number of bins ($D$)                         & $1000$ & $500$ \\
Minimum bin value ($b_0$)                    & $-5$   & $-5$ \\
Maximum bin value ($b_D$)                    & $5$    & $5$ \\
Number of channels ($K$)              & $C$    & $C$ \\
K-size, conv-2D  ($s_1$) & $3$ & $3$ \\
K-size, conv-1D ($s_2$) & $3$ & $3$ \\
K-size, final conv-1D ($s_3$)            & $51$ & $51$ \\
Number of blocks ($M$)               & $3$    & $3$ \\
Dropout                                      & $0.35$ & $0.35$ \\
Learning rate                                & $0.001$ & $0.001$ \\
\bottomrule
\end{tabular}
\end{table}

The comparison between models is presented in Table~\ref{tab:resutls}. We trained and evaluated all models using five different random seeds, and report the minimum, average, and maximum scores, as well as the standard deviation for each model. The results are summarized as follows:
\begin{itemize}
    \item For every dataset and baseline, the average score is worse than the BinConv average score plus one standard deviation, indicating consistently better performance by BinConv.
    \item Except for the weekly dataset where TimeGrad achieves the best minimum score, BinConv outperforms all baselines in terms of minimum, average, and maximum scores across all datasets.
    \item For the CRPS metric, the improvement achieved by BinConv is statistically significant (p-value for T-test $< 0.05$) for all datasets and baselines, except TimeGrad on the weekly dataset.
    \item For the NMAE metric, the improvement is statistically significant (p-value for T-test $< 0.05$) for all datasets and baselines, except TimeGrad on the weekly dataset and DLinear on the daily dataset.
\end{itemize}

Additionally, Table~\ref{tab:argmax_sampling} compares the performance of BinConv when using argmax-based versus sampling-based forecasting (see Section~\ref{sec:forecasting} for details). The results show that argmax-based forecasting does not degrade point forecasting performance.

\begin{table*}[ht]
\centering
\caption{CRPS and NMAE results for all models. Each metric is reported as the average (avg), minimum (min), maximum (max), and standard deviation (std) over 5 different seeds. An asterisk ($^*$) over the BinConv average score indicates that BinConv statistically significantly outperforms all baselines for that metric and dataset ($p < 0.05$).}
\label{tab:resutls}
\resizebox{\textwidth}{!}{
\begin{tabular}{llcccccccccc}
\toprule
\textbf{Dataset} & \textbf{Stat} &
\multicolumn{2}{c}{\textbf{DLinear}} &
\multicolumn{2}{c}{\textbf{PatchTST}} &
\multicolumn{2}{c}{\textbf{GRU-NVP}} &
\multicolumn{2}{c}{\textbf{TimeGrad}} &
\multicolumn{2}{c}{\textbf{BinConv}} \\
& &
\textbf{CRPS} & \textbf{NMAE} &
\textbf{CRPS} & \textbf{NMAE} &
\textbf{CRPS} & \textbf{NMAE} &
\textbf{CRPS} & \textbf{NMAE} &
\textbf{CRPS} & \textbf{NMAE} \\
\midrule
daily  & avg & 0.0440 & 0.0440 & \underline{0.0438} & \underline{0.0438} & 0.0714 & 0.0782 & 0.0587 & 0.0696 & \textbf{0.0327}$^{*}$ & \textbf{0.0382} \\
weekly & avg & 0.1046 & 0.1046 & 0.1039 & \underline{0.1039} & \underline{0.0972} & 0.1062 & 0.1046 & 0.1283 & \textbf{0.0902} & \textbf{0.0972} \\
tourism & avg & 0.2745 & 0.2745 & 0.2633 & \underline{0.2633} & 0.3723 & 0.4536 & \underline{0.2295} & 0.2968 & \textbf{0.1824}$^{*}$ & \textbf{0.1955}$^{*}$ \\
\midrule
daily  & min & \underline{0.0394} & \underline{0.0394} & 0.0404 & 0.0404 & 0.0570 & 0.0649 & 0.0402 & 0.0475 & \textbf{0.0291} & \textbf{0.0341} \\
weekly & min & 0.1007 & 0.1007 & 0.1020 & 0.1020 & 0.0916 & 0.1034 & \textbf{0.0800} & \underline{0.0985} & \underline{0.0851} & \textbf{0.0921} \\
tourism & min & 0.2595 & 0.2595 & 0.2281 & \underline{0.2281} & 0.2880 & 0.3849 & \underline{0.1993} & 0.2596 & \textbf{0.1777} & \textbf{0.1911} \\
\midrule
daily  & max & 0.0506 & 0.0506 & \underline{0.0476} & \underline{0.0476} & 0.1196 & 0.1259 & 0.0933 & 0.1077 & \textbf{0.0366} & \textbf{0.0427} \\
weekly & max & 0.1072 & \underline{0.1072} & 0.1078 & 0.1078 & \underline{0.1006} & 0.1085 & 0.1718 & 0.2121 & \textbf{0.0981} & \textbf{0.1055} \\
tourism & max & 0.3077 & 0.3077 & 0.2888 & \underline{0.2888} & 0.5799 & 0.5399 & \underline{0.2500} & 0.3186 & \textbf{0.1892} & \textbf{0.2012} \\
\midrule
daily  & std & 0.0043 & \underline{0.0043} & \textbf{0.0031} & \textbf{0.0031} & 0.0270 & 0.0266 & 0.0221 & 0.0252 & \underline{0.0037} & \underline{0.0043} \\
weekly & std & \underline{0.0025} & \underline{0.0025} & \textbf{0.0023} & \textbf{0.0023} & 0.0038 & 0.0026 & 0.0389 & 0.0488 & 0.0050 & 0.0052 \\
tourism & std & 0.0203 & \underline{0.0203} & 0.0227 & 0.0227 & 0.1216 & 0.0696 & \underline{0.0200} & 0.0235 & \textbf{0.0054} & \textbf{0.0049} \\
\bottomrule
\end{tabular}
}
\end{table*}

\begin{table}[ht]
\centering
\caption{Comparison of BinConv performance using sampling versus argmax forecasting. Each metric is reported as the average (avg), minimum (min), maximum (max), and standard deviation (std) over 5 different seeds.}
\label{tab:argmax_sampling}

\begin{tabular}{llcccc}
\toprule
\textbf{Dataset} & \textbf{Stat} &
\multicolumn{2}{c}{\textbf{Argmax}} &
\multicolumn{2}{c}{\textbf{Sampling}} \\
& &
\textbf{CRPS} & \textbf{NMAE} &
\textbf{CRPS} & \textbf{NMAE} \\
\midrule
daily         & avg & 0.037 & \textbf{0.037} & \textbf{0.033} & 0.038 \\
weekly        & avg & 0.098 & 0.098 & \textbf{0.090} & 0.098 \\
tourism  & avg & 0.196 & 0.196 & \textbf{0.182} & 0.196 \\
\midrule
daily         & min & 0.035 & 0.035 & \textbf{0.029} & \textbf{0.034} \\
weekly        & min & 0.093 & 0.093 & \textbf{0.085} & \textbf{0.092} \\
tourism  & min & 0.192 & 0.192 & \textbf{0.178} & \textbf{0.191} \\
\midrule
daily         & max & 0.039 & \textbf{0.039} & \textbf{0.037} & 0.044 \\
weekly        & max & 0.105 & \textbf{0.105} & \textbf{0.098} & 0.106 \\
tourism  & max & 0.201 & 0.201 & \textbf{0.189} & 0.201 \\
\midrule
daily         & std & \textbf{0.002} & 0.005 & 0.004 & \textbf{0.004} \\
weekly        & std & 0.005 & 0.005 & 0.005 & 0.005 \\
tourism  & std & 0.005 & 0.005 & 0.005 & 0.005 \\
\bottomrule
\end{tabular}%
\end{table}

\subsubsection{Multivariate Datasets}
\begin{table*}[ht]
\centering
\caption{
Forecasting performance (CRPS and NMAE) across different models and datasets. All datasets use context length 96 and prediction horizon 96 (except Illness, which uses 36 for both). Best scores per row are in bold. Bottom row shows average rank (lower is better).
}
\label{tab:model_comparison_lt}
\begin{tabular}{lccccccccccc}
\toprule
\textbf{Dataset} 
& \multicolumn{2}{c}{\textbf{DLinear}} 
& \multicolumn{2}{c}{\textbf{PatchTST}} 
& \multicolumn{2}{c}{\textbf{GRU-NVP}} 
& \multicolumn{2}{c}{\textbf{TimeGrad}} 
& \multicolumn{2}{c}{\textbf{BinConv}} \\
 & CRPS & NMAE & CRPS & NMAE & CRPS & NMAE & CRPS & NMAE & CRPS & NMAE \\
\midrule
ETTh1         & 0.3339 & 0.3339 & 0.3218 & 0.3218 & 0.3435 & 0.4383 & \textbf{0.2930} & 0.3867 & 0.2969 & \textbf{0.3201} \\
ETTh2         & 0.2319 & 0.2319 & 0.1763 & \textbf{0.1763} & 0.2623 & 0.3100 & 0.2059 & 0.2574 & \textbf{0.1709} & 0.1853 \\
ETTm1         & \textbf{0.2663} & \textbf{0.2663} & 0.2681 & 0.2681 & 0.4029 & 0.5018 & 0.4201 & 0.5153 & 0.2922 & 0.3141 \\
ETTm2         & 0.1431 & 0.1431 & 0.1337 & \textbf{0.1337} & 0.2106 & 0.2618 & 0.1595 & 0.1930 & \textbf{0.1329} & 0.1456 \\
Illness       & 0.1869 & 0.1869 & 0.1103 & 0.1103 & 0.0750 & \textbf{0.0828} & \textbf{0.0643} & 0.0868 & 0.1187 & 0.1447 \\
\midrule
\textbf{Avg. Rank} 
& 3.4 & 2.8 
& 2.4 & \textbf{1.8} 
& 4.2 & 4.0 
& 2.8 & 3.8 
& \textbf{2.2} & 2.6 \\
\bottomrule
\end{tabular}
\end{table*}

\label{app:eval_more_datasets}
Due to the high computational cost, particularly for the TimeGrad model, which requires up to 20 hours for training and evaluation, we ran a single seed per model and multivariate dataset. Each dataset contains 7 variables. Since BinConv currently supports only univariate forecasting, and extending it to the multivariate setting is left for future work, we forecasted each time series independently using the sampling procedure described in Section~\ref{sec:forecasting}.

In Table~\ref{tab:model_comparison_lt}, we report CRPS and NMAE for all baseline models and BinConv. Since there is no consistent winner across all datasets, we additionally report average model rankings for both metrics. Rankings were computed based on relative performance: the best-performing model on each dataset was assigned rank 1, the second-best rank 2, and so on.

Remarkably, despite using untuned hyperparameters and not being explicitly designed for multivariate forecasting, BinConv achieved the best average rank based on the CRPS metric and the second-best rank based on the NMAE metric.

\subsection{Efficiency of the Model}
In Table~\ref{tab:total-params}, we report the total number of parameters for each model across all datasets. The DLinear model has significantly fewer parameters than the others, with BinConv being the second most parameter-efficient.

In Table~\ref{tab:m4-daily-times}, we present the average computation time during both training and inference for all models on the M4 Daily dataset. To obtain these results, we trained and evaluated all models using a batch size of $1$, and computed the mean and standard deviation over all samples and epochs. As expected, DLinear is the fastest model during training, with BinConv being the second fastest.

During inference, PatchTST is the fastest model because it is non-autoregressive and predicts the entire forecast horizon in a single forward pass. DLinear ranks second, followed by BinConv and GRU-NVP, which exhibit similar inference speeds--approximately twice as slow as PatchTST. TimeGrad is significantly slower than all other models.

Using point (argmax-based) forecasting instead of sampling improves inference speed by approximately 25\%. However, the difference is not dramatic, as sampling is implemented by duplicating the input $N$ times, where $N$ is the number of samples. This allows all $N$ samples to be processed in a single batch.

\begin{table*}[ht]
\centering
\caption{Total number of parameters (in thousands) per method and dataset.}
\label{tab:total-params}
\begin{tabular}{lrrrrr}
\toprule
\textbf{Dataset} & \textbf{DLinear} & \textbf{BinConv} & \textbf{PatchTST} & \textbf{TimeGrad} & \textbf{GRU-NVP} \\
\midrule
daily         & \textbf{1.204}   & 20.173 & 47.470   & 107.434  & 141.108 \\
weekly        & \textbf{1.040}   & 17.680 & 43.725   & 107.434  & 141.108 \\
tourism  & \textbf{3.504}   & 54.013 & 155.288  & 107.434  & 141.108 \\
\bottomrule
\end{tabular}

\end{table*}

\begin{table*}[ht]
\centering
\caption{Mean and standard deviation of time (in seconds) to process a single sample for the M4 Daily dataset.}
\label{tab:m4-daily-times}
\resizebox{\textwidth}{!}{%
\begin{tabular}{lcccccc}
\toprule
\textbf{Phase} & \textbf{DLinear} & \multicolumn{2}{c}{\textbf{BinConv}} & \textbf{PatchTST} & \textbf{TimeGrad} & \textbf{GRU-NVP} \\
               &                  & Argmax & Sampling                      &                   &                   &                   \\
Train     & $\mathbf{0.0123 \pm 0.0026}$ & $0.0284 \pm 0.0043$ & $0.0284 \pm 0.0043$  & $0.0381 \pm 0.0049$ & $0.0450 \pm 0.0063$ & $0.0362 \pm 0.0048$ \\
Inference      & $0.0755 \pm 0.0016$ & $0.09684 \pm 0.00306$ & $0.12788 \pm 0.0017$ & $\mathbf{0.0547 \pm 0.0021}$ & $2.7723 \pm 0.0365$ & $0.1239 \pm 0.0086$ \\
\bottomrule
\end{tabular}%
}
\end{table*}

\section{Ablation Study}
\label{sec:ablation_study}

\subsection{Fully Connected Layers with BinConv}
\label{sec:fc_binconv}

When designing the BinConv architecture, we intentionally excluded fully connected (FC) layers to maintain both strong generalization and parameter efficiency. Compared to convolutional layers, FC layers introduce significantly more trainable parameters, particularly problematic in high-dimensional output spaces. More critically, due to the design of CBE and the associated loss computation, an FC layer can restrict the model’s ability to forecast values beyond those observed during training. This limitation arises because FC layers lack weight sharing, meaning each output component is governed by an independent set of parameters. Consequently, if the last $K$ elements of a CBE target vector are frequently zero in the training set (a plausible scenario given that most targets are well below the maximum bin value $b_D$), the model may trivially learn to output zeros for these components. As a result, such an architecture fails to extrapolate to higher values during testing.

\begin{figure}[ht]
    \includegraphics[width=\columnwidth]{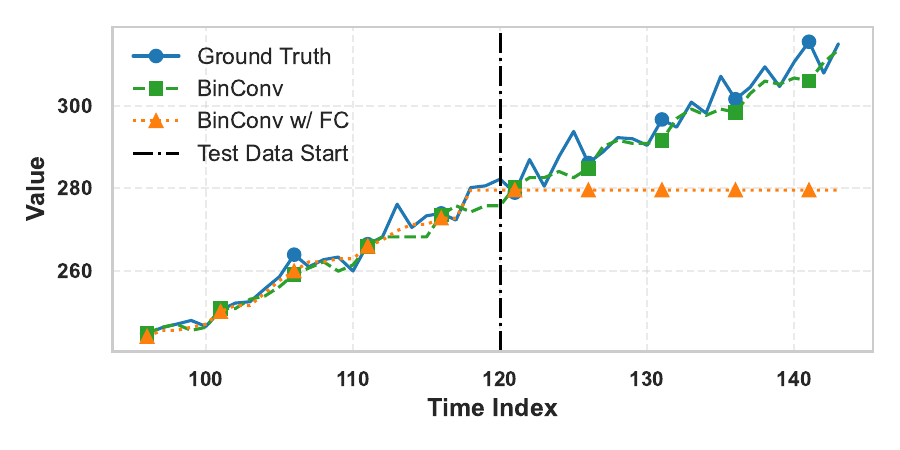}
    \caption{Forecast comparison of BinConv and BinConv w/ FC on synthetic linear trend data. The vertical line marks the start of the test data. BinConv w/ FC fails to forecast values beyond those seen during training.}
    \label{fig:forecast_comparison}
\end{figure}

To illustrate this, we implemented a BinConv variant in which the final 1D convolution was replaced by an FC layer. The output of the BinConv blocks, $h \in \mathbb{R}^{C \times D}$, where $C$ is the context length and $D$ the number of bins, was averaged over the context dimension to yield $\bar{h} \in \mathbb{R}^D$, which was then mapped to $\mathbb{R}^D$ via the FC layer. Using a synthetic dataset with a linear trend,
\[
s_t = (100 + 1.5\,t)(1 + \sigma_t), \quad \sigma_t \sim \mathcal{N}(0,10^{-4}),
\]
We compared forecasts from the original BinConv and its FC-layer variant. While the original model successfully extrapolated the trend, the FC variant saturated at the maximum value seen during training, failing to capture future growth (Figure~\ref{fig:forecast_comparison}). To highlight this issue, we used per-dataset mean scaling on a synthetic example. Per-sample mean scaling partially mitigates the problem, as higher test-time values shift the mean, enabling some extrapolation.

Table~\ref{tab:ablation_metrics} reports evaluation metrics for the FC variant with per-sample scaling (column \textquotedblleft w/ FC Layer\textquotedblright). Despite reasonable results, the model cannot produce CBE vectors with more active bins than seen during training, as many neurons in the final layer become inactive and output constant values. In addition, as shown in Table~\ref{tab:ablation_params}, the FC layer significantly increases the number of parameters. Overall, the FC variant remains ill-suited for real-world forecasting tasks.

\subsection{Transformer with CBE}

\iffalse
Instead of replacing only the final layer with a fully connected layer, one might consider replacing the entire BinConv architecture with a transformer-based model, which relies exclusively on fully connected layers. However, in practice, this substitution leads to a substantial drop in performance. We experimented with both encoder-decoder and decoder-only transformer architectures, and in both cases, the results were poor.

For instance, on the weekly dataset, the NMAE increased by an order of magnitude compared to BinConv, reaching up to $0.8$ in our best attempt. We attribute this performance degradation to the nature of transformer architectures, which use different weights to process each input component and to generate each output component. While this is appropriate for one-hot encoding, where the model is exposed to all possible one-hot vectors during training, it is not suitable for CBE.

As discussed in Section~\ref{sec:fc_binconv}, it is highly unlikely that a CBE representation encountered during training will contain cases where almost all components are equal to one or zero. This structural property makes CBE better suited to architectures with shared weights across components, where the model's behavior does not depend on the position of a component in the vector. Convolutional architectures naturally satisfy these requirements, making them a more appropriate choice for processing CBE representations.
\fi

An alternative to modifying only the final layer of BinConv is to replace the entire architecture with a transformer-based model, which relies exclusively on FC layers. However, our experiments show that this substitution results in a substantial performance drop. We evaluated both encoder–decoder and decoder-only transformer variants, and in all cases the results were unsatisfactory. For example, on the weekly dataset the NMAE increased by nearly an order of magnitude compared to BinConv, reaching values as high as $0.8$ in the best-performing transformer configuration.

We attribute this degradation to the fundamental design of transformer architectures, which employ distinct weight matrices for processing each input component and generating each output component. While such flexibility is appropriate for one-hot encodings, it is ill-suited for the continuous bin encoding used in BinConv. As discussed in Section~\ref{sec:fc_binconv}, CBE representations rarely contain vectors in which most components are exactly one or zero. This structural property favors architectures with weight sharing across components, ensuring that the model’s behavior is invariant to component position. Convolutional architectures naturally satisfy these constraints, making them more appropriate than transformers for processing CBE representations.

\subsection{One-Hot Encoding with BinConv}

In this section, we show that BinConv woorks poorly with one-hot encoding. We modify BinConv in a following way: we replace CBE with one-hot encoding both for input and output; replace final activation fucntion from sigmoid to softmax and utilze standard multi-class cross entropy loss. For this variation NMAE on the weekly dataset increases from 0.09 to 1.18, which is notably worse than all other models and baselines considered in this study. The explanation of that is that BinConv architecture, as most convolution architectures, relies heavily on neighbours component, when it forecast the   CBE is replaced with one-hot encoding, the convolutional network struggles to extract meaningful features because most of the receptive field is filled with zeros. This severely limits the network's learning capacity, resulting in significantly degraded performance. For example, NMAE on the weekly dataset increases from 0.09 to 1.18, which is notably worse than all other models and baselines considered in this study.

Unlike fully connected architectures such as transformers, BinConv operates over small local windows and therefore relies heavily on the relationships between neighboring components in the input encoding. This locality makes convolution particularly effective for structured encodings like CBE, where adjacent components often carry correlated information. 

In contrast, for one-hot encodings, where exactly one component is equal to one and all others are zero, there is no meaningful local structure to exploit. The neighbors of the component with value one are always zero by design, making the surrounding values uninformative. In CBE, however, if a given component is equal to one, its neighboring components may still be either one or zero. For example, components to the right of a one-valued component may also be one, and components to the left of a zero-valued component may still be one. This structure enables convolutional filters to leverage informative local patterns, which is not possible with one-hot encodings.

\begin{table}[ht]
\centering
\caption{Comparison of architectural variants against the standard BinConv in terms of NMAE. The notation \textbf{w/} indicates that the standard BinConv is modified by replacing a specific component: \textbf{w/FC} uses a fully connected layer instead of the final convolution, \textbf{w/SC} replaces Depthwise convolutions with standard ones, colored arrows indicate whether a variant performs better (\textcolor{green!50!black}{↓}) or worse (\textcolor{red}{↑}) than BinConv.}
\label{tab:ablation_metrics}
%\begin{tabular}{lcccccc}
%\toprule
%\textbf{Dataset} & \multicolumn{2}{c}{\textbf{BinConv}} & \multicolumn{2}{c}{\textbf{w/FC}} & \multicolumn{2}{c}{\textbf{w/SC}} \\
% & CRPS & NMAE & CRPS & NMAE & CRPS & NMAE \\
%\midrule
%daily         & 0.0327 & 0.0382 & 0.0411\,\textcolor{red}{↑} & 0.0468\,\textcolor{red}{↑} & 0.0292\,\textcolor{green!50!black}{↓} & 0.0343\,\textcolor{green!50!black}{↓} \\
%weekly        & 0.0902 & 0.0972 & 0.0870\,\textcolor{green!50!black}{↓} & 0.0943\,\textcolor{green!50!black}{↓} & 0.1028\,\textcolor{red}{↑} & 0.1112\,\textcolor{red}{↑} \\
%tourism  & 0.1824 & 0.1955 & 0.1756\,\textcolor{green!50!black}{↓} & 0.1910\,\textcolor{green!50!black}{↓} & 0.1801\,\textcolor{green!50!black}{↓} & 0.1945\,\textcolor{green!50!black}{↓} \\
%\bottomrule
%\end{tabular}
\begin{tabular}{lccc}
\toprule
\textbf{Dataset} & \textbf{BinConv} & \textbf{w/FC} & \textbf{w/SC} \\
\midrule
daily & 0.0382 & 0.0468\,\textcolor{red}{↑} & 0.0343\,\textcolor{green!50!black}{↓} \\
weekly & 0.0972 & 0.0943\,\textcolor{green!50!black}{↓} & 0.1112\,\textcolor{red}{↑} \\
tourism & 0.1955 & 0.1910\,\textcolor{green!50!black}{↓} & 0.1945\,\textcolor{green!50!black}{↓} \\
\bottomrule
\end{tabular}
\end{table}

\begin{table}[ht]
\centering
\caption{
Number of parameters (in thousands) for the standard BinConv model and its architectural variations. 
See the caption of Table~\ref{tab:ablation_metrics} for the explanation of the column notation.
}
\label{tab:ablation_params}
\begin{tabular}{lccc}
\toprule
\textbf{Dataset} & \textbf{BinConv} & \textbf{w/FC} & \textbf{w/SC} \\
\midrule
daily         & \textbf{20.173} & 1019.540 & 51.679 \\
weekly        & \textbf{17.680} & 1017.164 & 44.830 \\
tourism  & \textbf{54.013} & 1052.210 & 146.899 \\
\bottomrule
\end{tabular}
\end{table}

\subsection{Depthwise Convolution instead of Standard Convolution}
\label{sec:dwconv}

Depthwise convolution is a form of convolution in which each input channel is convolved independently with its own filter. As a result, it requires significantly fewer convolutional kernels, which improves the model's computational efficiency. In our model, we use this type of convolution in the BinConv block for 1D operations.

The performance of BinConv with standard 1D convolutional layers in place of depthwise layers is reported in Table~\ref{tab:ablation_metrics}, under the column labeled \textquotedblleft w/SC\textquotedblright. Replacing depthwise convolutions with standard ones does not improve performance on any of the datasets, and it reduces model efficiency, as shown in Table~\ref{tab:ablation_params}.

\section{Conclusion}

In this paper, we proposed using Cumulative Binary Encoding (CBE) as a discrete quantization technique for time series, and we introduced BinConv, a convolutional architecture specifically designed for efficient processing of CBE vectors. Extensive evaluation on the publicly available ProbTS benchmark demonstrates that our method achieves state-of-the-art performance on univariate time series datasets and performs no worse on multivariate ones compared to strong baseline models. Through a comprehensive ablation study, we show that neither CBE with fully connected architectures nor BinConv with one-hot encoding performs nearly as well as the combination of CBE and BinConv.

Despite these promising results, our method has several limitations. First, CBE discretizes time series values, which requires approximating the precision of continuous data and can result in information loss. Second, selecting the number of bins introduces a trade-off: too few bins reduce representational fidelity, while too many increase computational complexity. Third, the current design is limited to univariate time series, which necessitates modeling each variable independently and prevents capturing cross-series interaction.

Future directions for this work include evaluating the generalization capability of our method in a Time Series Foundation Model setting, where the model is tested on previously unseen datasets. Another promising avenue is to extend the BinConv architecture to support multivariate time series by modeling interactions between components. Additionally, our approach could be applied to other time series tasks, such as imputation, anomaly detection, and classification.

\bibliography{aaai2026}

\end{document}